# ARiA: Utilizing Richard's Curve for Controlling the Non-monotonicity of the Activation Function in Deep Neural Nets

**Narendra Patwardhan**
Michigan Technological University
narendra@mtu.edu

**Madhura Ingalhalikar**
Symbiosis Centre for Medical Image Analysis
madhura.ingalhalikar@sitpune.edu.in

**Rahee Walambe**
Symbiosis Institute of Technology
rahee.walambe@sitpune.edu.in

## Abstract

This work introduces a novel activation unit that can be efficiently employed in deep neural nets (DNNs) and performs significantly better than the traditional Rectified Linear Units (ReLU). The function developed is a two parameter version of the specialized Richard's Curve and we call it Adaptive Richard's Curve weighted Activation (ARiA). This function is non-monotonous, analogous to the newly introduced Swish, however allows a precise control over its non-monotonous convexity by varying the hyper-parameters. We first demonstrate the mathematical significance of the two parameter ARiA followed by its application to benchmark problems such as MNIST, CIFAR-10 and CIFAR-100, where we compare the performance with ReLU and Swish units. Our results illustrate a significantly superior performance on all these datasets, making ARiA a potential replacement for ReLU and other activations in DNNs.

## 1 Introduction

The choice of activation function $f(.)$ in deep neural nets is critical as it not only controls the firing of the neurons but also influences the training dynamics as well as directly affects the prediction accuracies[2]. To this date, rectified linear units (ReLU)[15] defined by $f(x) = x^+ = max(0, x)$, have been a popular choice, especially in convolutional neural nets (CNNs) because of their simplicity, monotonicity and gradient preserving property as opposed to Tanh and/or Sigmoid activations that are cursed by the vanishing gradient problem[8]. However, ReLU activations profoundly suffer from the dying neuron problem as significant percent of the neurons get permanently deactivated if the pre-activation is $\leq 0$ as ReLU regularizes the gradient to zero. To resolve this problem, numerous modifications to ReLU have been proposed and include the leaky ReLU[14], Max-out[9], ELU[4][19] etc. These variations, albeit, are non-zero for $x < 0$, are monotonous and are unbounded to the region of significance, culminating into inconsistencies in their performance.

Google Brain recently introduced a smooth and non-monotonous function based off the sigmoid function named Swish [17] defined by $x * \sigma(\beta, x)$. It uses a single hyper-parameter $\beta$ which not only controls the slope of the function but also the rate at which the first derivative asymptotes to 0 and 1. The non-monotonous curvature present in the region of significance of Swish can be controlled by varying the value of $\beta$ through which the network can be optimized. Here, we define the region of significance as the small area around the origin on x-axis [17][8][6]. Consistent superior performance of Swish has been demonstrated on a number of benchmark classification problems as shown in

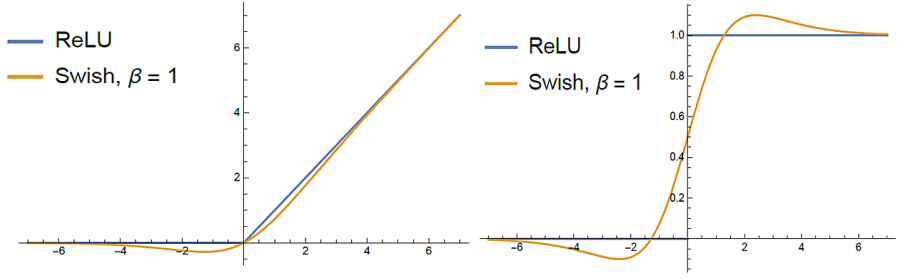

Figure 1: (a)ReLU and Swish Functions (b)Derivative of ReLU and Swish

Ramachandran et.al.[17].

Despite such remarkable performance, Swish is limited in its application as the hyper-parameter $\beta$ cannot facilitate precise control over the non-monotonous convexity of the function as a small change in $\beta$ affects the Swish curve in both the positive and negative regions.

To this end, we introduce a novel activation function defined by $x * \sigma(L)$ which is a function derived from the Richard's Curve and we call it as Adaptive Richard's Curve weighted Activation (ARiA). In this work, we focus primarily on a two parameter adaptation of ARiA that facilitates accurate control over the non-monotonous convexity of the function, independently, in both the first and third quadrants. We first discuss the mathematical premise for ARiA, its two parameter adaptation (ARiA2) and subsequently validate its performance on the standard benchmark problems namely MNIST, CIFAR-10 and CIFAR-100 data sets where we employ distinct neural net architectures. We compare performance of ARiA2 with Swish and ReLU activations for these datasets and demonstrate that ARiA not only exhibits faster learning but also outperforms its counterparts.

## 2 Methods

In this section, we introduce our novel activation function named 'ARiA' and demonstrate its mathematical relevance in comparison to other activation functions that include Swish and ReLU. We begin with recapitulating ReLU and Swish in brief and then describe the new function.

### 2.1 ReLU

ReLU is defined as

$$f(x) = x^+ = max(0, x) \quad (1)$$

where, $x$ is the input to the neuron. Conventionally, ReLU has been the most popular activation function due to its gradient preserving property (i.e. having a derivative of 1 for $x > 0$) as opposed to Tanh which typically suffers from vanishing gradient problem [10]. By definition, ReLU is a monotonous and a smooth function as shown in Fig. 1(a) and (b)). However, for randomly initialized pre-activations, the networks extensively suffers from the dying neuron problem where a high percentage of neurons are deactivated, subsequently deteriorating the network efficiency[4].

### 2.2 Swish

The standard logistic sigmoid function is defined by

$$\sigma(x) = (1 + e^{-x})^{-1} \quad (2)$$

Swish function is an adaptation of logistic sigmoid function given in Eq.(2) obtained by multiplying Eq.(2)with raw pre-activation. It introduces a hyper-parameter $\beta$ and is defined by Eq.(3).

$$f(x) = x * \sigma(\beta, x) \quad (3)$$

where

$$\sigma(\beta, x) = (1 + e^{-\beta x})^{-1} \quad (4)$$

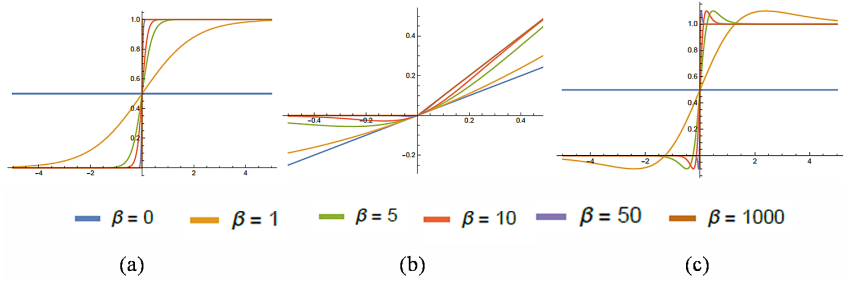

Figure 2: (a)Variation in Logistic Sigmoid function with varying $\beta$(b)Variation in Swish with change in $\beta$(c)Derivative of function in (b)
.

Fig. 2(a) shows the plots for logistic sigmoid function given in Eq.(4) obtained by varying $\beta$. It can be observed that the function varies between heaviside step function (for extremely large values of $\beta = 1000+$) and a scaled linear function (for $\beta = 0$). Fig. 2(b) displays the Swish function given in Eq.(3) for various values of $\beta$ where, for $\beta \to \infty$, the Swish function approaches ReLU and at $\beta = 0$, it becomes a scaled linear function $f(x) = \frac{x}{2}$. Finally, for $\beta = 1$, Swish is equivalent to the Sigmoid-weighted linear unit (SiL)[5].

Next, we plot the derivatives for the Swish function as shown in Fig. 2(c). The scale of $\beta$ controls the rate at which the first derivative approaches 0 and 1. Swish differs from ReLU in mainly two aspects: (1) the 'non-monotonic curvature' or convex curvature for $x < 0$ in third quadrant, mainly in the pre-activation region of the function as shown in Fig. 2(b) and (2) the smoother slopes for the linear part of the derivative plots in the first quadrant. From Fig. 2(c), we can infer that, with change in the value of $\beta$, the curvature in the third quadrant can be coarsely controlled, however, this also modulates the slope of the function in the first quadrant. For example, if $\beta$ is increased, the curvature in the third quadrant reduces, controlling the $f(.)$ for $x < 0$, however, the slope in the first quadrant increases that may subsequently increase the $f(.)$ value through the multiple layers of the neural net leading to an exploding gradient[16].This limits the application of the Swish function, as it cannot control the function in both quadrants independently.

## 2.3 ARiA

We introduce a novel activation function that is based on a form of Richard's Curve[18] and we abbreviate it as ARiA. Richard's curve is a generalized logistic function that allows more flexibility in its S-shape. The sigmoid or the logistic functions are limited as these are rotationally symmetric around the point of inflection. Therefore, to generalize these for cases where rotational asymmetry is important, Richard's [18] introduced a multi-parameter function as defined in the Eq.(5).

$$\sigma(L) = f(x, A, K, B, \nu, Q, C) = A + \frac{K - A}{(C + Qe^{-Bx})^{\frac{1}{\nu}}} \quad (5)$$

where, $A$ is the lower asymptote; $K$ is the upper asymptote; $B$ is the exponential growth rate; $\nu > 0$ decides the direction of growth while $Q$ is related to the initial value of function $\sigma(L)$ and $C$ is a constant which typically is chosen as one [7]. Based on this, we define our novel function ARiA, as the product of raw pre-activations and Richard's Curve and is give in Eq.(6)

$$ARiA = x * \sigma(L) \quad (6)$$

The ARiA as well as Richard's Curve contain five hyper-parameters that affect the shape of the function differently. We demonstrate the importance of each parameter in Fig. 3. Fig. 3(a-h) display Richard's Curve and the ARiA function for change in every parameter respectively while $C$ is retained at 1. In Fig. 3(a & b), we vary parameter $Q$ and observe that it affects only the non-monotonic curvature. However, this curvature does not asymptote to zero rapidly as seen by the orange line at $Q = 2$ in Fig. 3(b). Next, we study parameter $B$ as shown in Fig. 3(c & d). This parameter is important as it introduces exponential non-linearity to the function (Fig. 3(c))and for ARiA, with increase in $B$ the non-monotonic curvature asymptotes quickly to zero as can be observed from

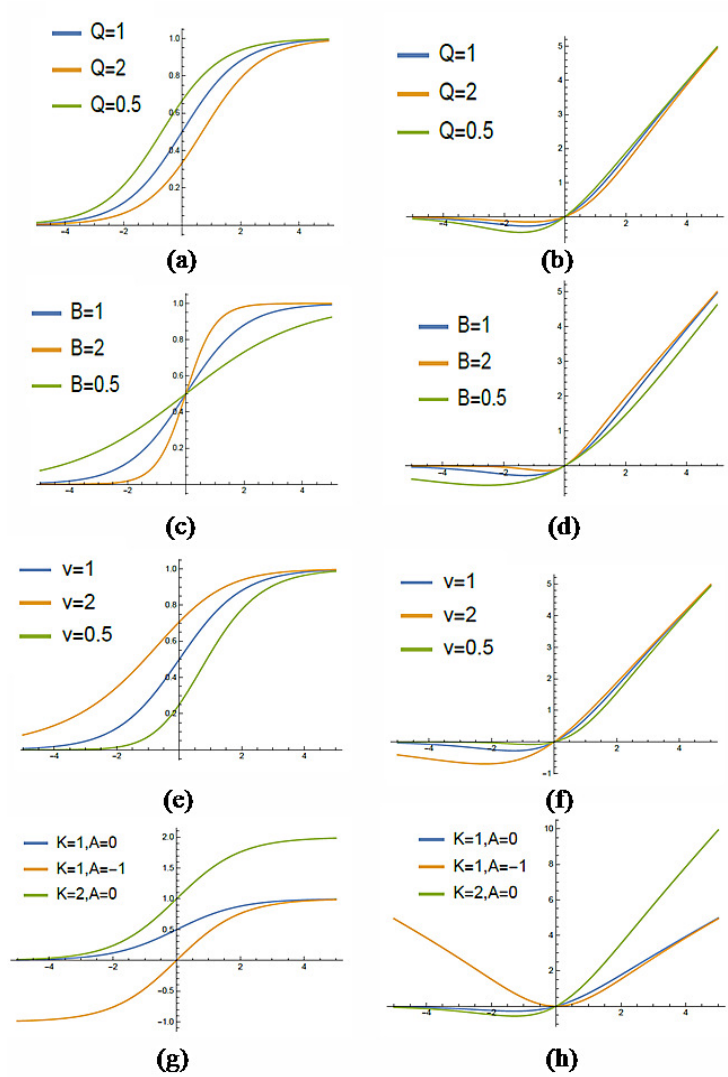

Figure 3: Variation of Parameters on Richard's Curve and ARiA Function

Fig. 3(d). The role of parameter $\nu$, is crucial as lower values of $\nu < 1$, demonstrate a lower gradient in the first quadrant while in the third quadrant these display a similar behavior as $B$ (Fig. 3(f)). Finally, changing $K$ scales the whole curve whereas simultaneously varying both $K$ & $A$ has much complex effect on the shape of both Richard's as well as ARiA as shown in Fig. 3(g & h).

Overall, all the parameters described above, have the capability to control the shape of ARiA and therefore if ARiA is employed as an activation function and is optimized to a particular problem, it it has the potential to boost the network performance. However, optimizing 4 to 5 parameters heavily increases the number of operations per layer consequently increasing the computational cost. Hence, it is critical to restrict the number of hyper-parameters and we achieve this by selecting the most influential variables in the region of significance. Based on the plots in Fig. 3, these include only two hyper-parameters $B$ (Fig. 3(d)) and $\nu$(Fig. 3(f)), which individually as well as in combination affect the curve in the region of significance and for higher values on both +ve and -ve sides, the curve remains unaltered. This reduces the ARiA function to Eq.(7), where we call these parameters $\alpha$ and $\beta$ where $\alpha = \frac{1}{\nu}$ and $\beta = B$. Henceforth, we call this two-parameter adaptation of ARiA as ARiA2.

$$f(x) = x * \sigma(\alpha, \beta, x) \tag{7}$$

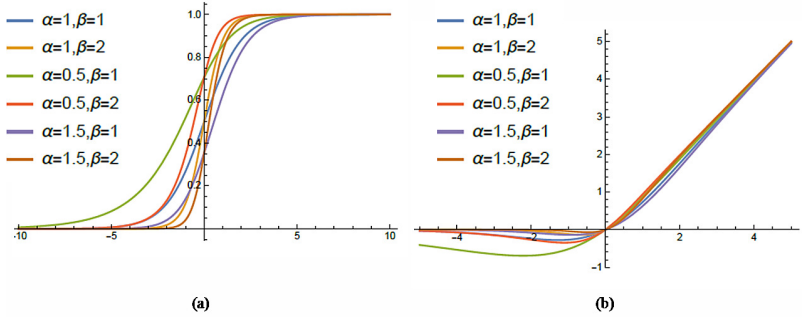

Figure 4: (a) Logistic Sigmoid curves by varying $\alpha$ and $\beta$ (b)ARiA2 by varying $\alpha$ and $\beta$ both

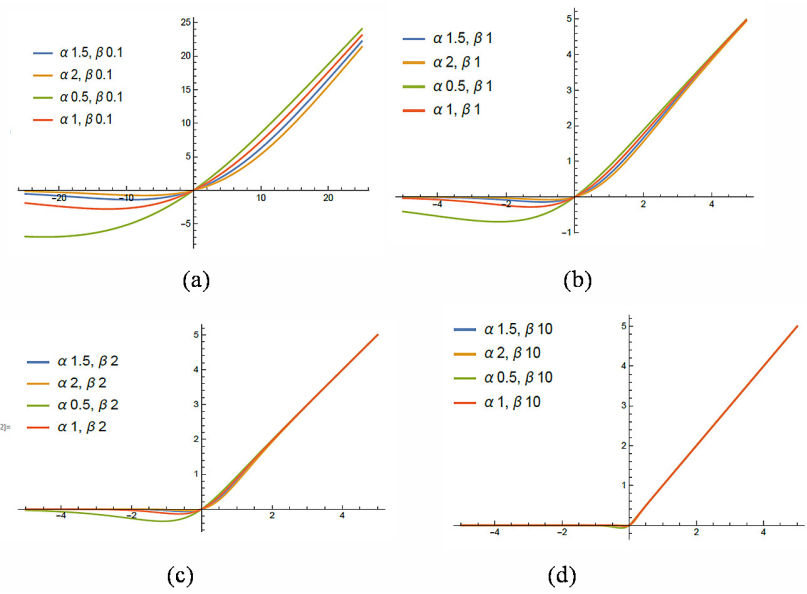

Figure 5: Function Plots for ARiA2

where
$$\sigma(\alpha, \beta, x) = (1 + e^{-\beta x})^{-\alpha} \qquad (8)$$

Comparing Eq.(3) of Swish and Eq.(7), it can be observed that by employing two hyper-parameters $\alpha$ and $\beta$ in combination, ARiA2 allows for a more complex control over the nature and shape of the curve facilitating better adaptation to a particular problem as compared to Swish and this has been described in the following section.

### 2.4 Comparison with Swish

Swish is a special case of ARiA, manifested at ARiA=$f(x, 1, 0, 1, 1, \beta, 1)$ whereas, ARiA2 is defined at ARiA=$f(x, 1, 0, 1, 1, 0, \beta, \frac{1}{\alpha})$. Fig. 4(a) shows the effect of varying both $\alpha$ and $\beta$ on the logistic sigmoid function and Fig. 4(b) shows the corresponding ARiA2 function. By comparing Fig. 2(b) and Fig. 4(b), it can be observed that an additional hyper-parameter facilitates superior control over the shape of the curve subsequently offering higher adaptability to the learning process.

Fig. 5(a-d) plots the graph of ARiA2 for different values of $\alpha = (0.5, 1.0, 1.5, 2.0)$ for $\beta = 0.1$(Fig. 5-a), $\beta = 1$(Fig. 5-b), $\beta = 2$(Fig. 5-c), $\beta = 10$(Fig. 5-d). The introduction of hyper-parameter $\alpha$ has a two-fold effect; it reduces the curvature in 3rd quadrant as well as increases the curvature in first quadrant while lowering the value of activation, for $\alpha > 1$. Overall, ARiA2 provides an accurate control over the non-monotonic convex curvature in the 3rd quadrant as shown in Fig. 5.

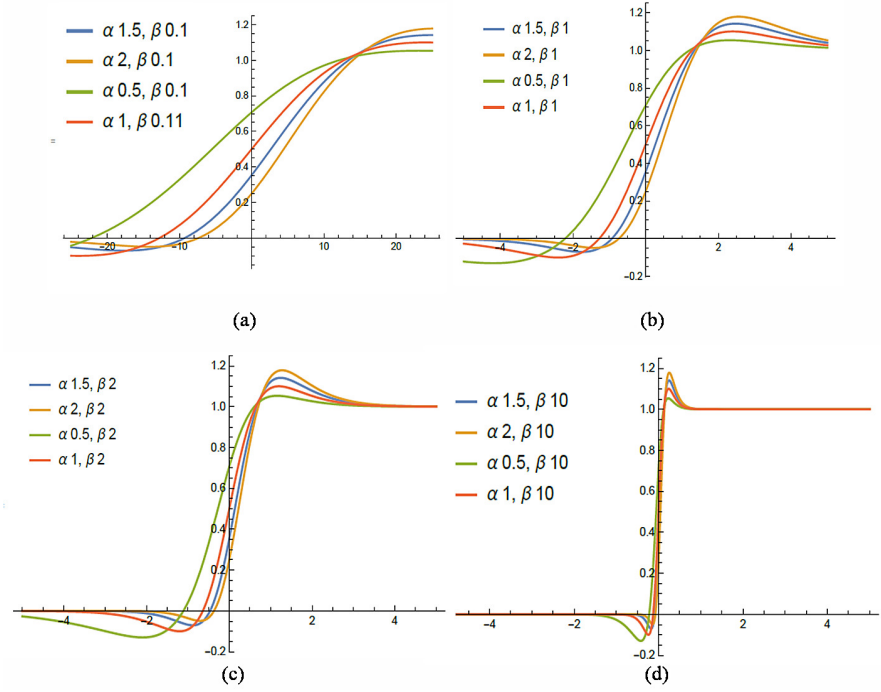

Figure 6: Derivative of Function Plots for ARiA2

The derivative of ARiA2 is given by:

$$A\dot{RiA}2 = (1 + e^{-\beta x})^{-\alpha} + e^{-\beta x}(1 + e^{-\beta x})^{-1-\alpha} x \alpha \beta \log[e] \tag{9}$$

The manipulation of $\alpha$ has more distinct effect than $\beta$ on the function shape and its derivative as seen in Fig. 6. Fig. 6(a-d) show the first derivative plots for ARiA2 function as displayed in Fig. 5(a-d). The analysis can be extended further by looking at the plots in Fig. 5 and Fig. 6. Let the activation function be defined as $f(x)$. The function plot can be subdivided into four regions of interest namely, large positive, small positive, small negative, large negative values of $x$.

### 2.4.1 Observations

- For small values of $x$ (positive and negative), ARiA2 (and Swish) exhibit a convex upside opening curvature which is completely absent in ReLU (Fig. 1). This lowers the activation value when small negative inputs are present, unlike ReLU for which such inputs do not contribute to the activation value. For small positive weights because of the non-linearity of Swish and ARiA2, the activation values are lower than ReLU. As can be observed from Fig. 5(b), the curve for ARiA2 ($\alpha = 0.5, \beta = 1$) has lower convexity in the 1st quadrant and higher convexity in the third quadrant as compared to the curves for $\alpha > 1$. Hence ARiA2 with $\alpha < 1$, usually underperforms Swish and ReLU.

- The ARiA2 curves for $\alpha > 1$, demonstrate a smaller curvature in the third quadrant and higher curvature in the first quadrant, more specifically, in the region of significance in comparison with Swish and ReLU functions. When such a function is employed in DNN architectures, the batch normalization layer[12] scales down the inputs to the next layer where the weights may perhaps are defined in the region of significance.

- Fig. 6 demonstrates that for large negative values, unlike ReLU, there exists a small gradient for ARiA2 (and Swish), while for large positive values the nature of the curve is similar to ReLU. The derivative of ARiA2 and Swish is continuous which may offer small computational advantage over ReLU whose derivative is undefined at 0 and needs to be defined piecewise.

Table 1: Testing Accuracy for MNIST using ReLU, Swish and ARiA2 with Custom CNN

| Activation Function | Testing Accuracy(%) |
|---|---|
| ReLU | 96.16 |
| ARiA2 ($\alpha = 1, \beta = 1$) | 96.35 |
| ARiA2 ($\alpha = 0.5, \beta = 1$) | 95.92 |
| ARiA2 ($\alpha = 0.75, \beta = 1$) | 96.35 |
| ARiA2 ($\alpha = 1.25, \beta = 1$) | 96.55 |
| ARiA2 ($\alpha = 1.5, \beta = 1$) | **96.60** |
| ARiA2 ($\alpha = 1.75, \beta = 1$) | 96.57 |
| ARiA2 ($\alpha = 2, \beta = 1$) | 96.54 |
| ARiA2 ($\alpha = 1.5, \beta = 2$) | **96.7** |

- As stated earlier, the control on the convexity in the small positive and small negative regions is of utmost importance for the activation function to provide more precision and stability to the model. Therefore, when comparing between ARiA2 and Swish functions, ARiA2 offers this capability via two hyper-parameters while Swish only performs coarse modulation through varying $\beta$ and this phenomenon can be observed through the derivative plots displayed in Fig. 6 where for ($\alpha = 1, \beta = 1$), ARiA2 reduces to the base definition of Swish.

## 3 Experiments and Results

The following sections first describe our experiments and the obtained results in greater detail. We have incorporated appropriate architectures pertaining to each of the dataset to illustrate how ARiA2 can perform on a wide range of neural net architectures. It is important to note that due to differences in training setup and computational capacity, our results may not be directly comparable to the results reported in the analogous works.

### 3.1 Datasets

#### 3.1.1 MNIST

The MNIST database (Modified National Institute of Standards and Technology database) is a large database of handwritten digits that is commonly used for training various image processing systems. The MNIST database contains 60,000 training images and 10,000 testing images. We tested the MNIST dataset using two separate architectures; 1)user defined CNN and 2) Dense Convolutional Network (DenseNet). For training of both of these models, AWS P2-xlarge instance with Keras having TensorFlow[1] backend was employed.
For the custom CNN, two blocks of convolutional layers were employed. Each block was followed by a pooling layer. The $2^{nd}$ pooling layer was followed by a dense layer consisting of 1024 neurons with dropout rate of 0.4, followed by $2^{nd}$ dense layer with 10 neurons, one for each target class (ranging from 0 to 9). Adam optimizer[13] with learning rate 0.001 was employed and the model was run for 30 epochs. Table 1 shows the testing accuracy for the MNIST dataset with this CNN model using ReLU, Swish and ARiA2 activation functions.

From Table 1, it can be seen that ARiA2 at $\alpha = 1.5, \beta = 2$, provides the highest accuracy. Moreover, the accuracy of ARiA2 at $\alpha = 1.5, \beta = 1$ also exceeds the ReLU and Swish results.

Next, we tested the MNIST dataset with state-of-art DenseNet to validate its usability in various architectures. It has been shown that DenseNet is an optimal model and requires less computational time to achieve better performance[11]. To train this model we employed Adam Optimizer with learning rate 0.0001 for 25 epochs. 2 cases of ARiA2 demonstrate better testing accuracy as shown in Table 2.

#### 3.1.2 CIFAR-10 and CIFAR-100

The CIFAR-10 is a labeled subset of 60000 32x32 color images of 10 classes, with 6000 images per class. There are 50000 training images and 10000 test images whereas CIFAR-100 is the labeled dataset 100 image classes, with 600 images per class. Each image is 32x32x3 (3 color), and the 600 images are divided into 500 training, and 100 test for each class. We train and validate the Wide Residual Network described by Zagoruyko et.al. [20] for CIFAR-10 & CIFAR-100. The CNN employed follows the original implementation by Zagoruyko et.al.[20] with dropouts. We train for 100 epochs with slightly higher initial learning rate of 0.125 with SGD learning algorithm[3] with a scheduled decay of 0.2. the experiments were performed using Keras with TensorFlow backend on an AWS P2-xlarge instance for CIFAR-10 and P2-8xlarge instance for CIFAR-100. Table 3 shows the prediction results using all three activation functions for CIFAR-10 and Table 4 shows the prediction results using all three activation functions for CIFAR-100.

Table 2: Testing Accuracy for MNIST using ReLU, Swish and ARiA2 using DenseNet

| Activation Function | Testing Accuracy(%) |
|---|---|
| ReLU | 98.72 |
| ARiA2 ($\alpha = 1, \beta = 1$) | 98.24 |
| ARiA2 ($\alpha = 0.5, \beta = 1$) | 98.29 |
| ARiA2 ($\alpha = 1.25, \beta = 1$) | **98.84** |
| ARiA2 ($\alpha = 1.5, \beta = 1$) | 98.74 |
| ARiA2 ($\alpha = 1.75, \beta = 1$) | **98.84** |
| ARiA2 ($\alpha = 2, \beta = 1$) | 98.33 |

Table 3: Testing Accuracy for CIFAR-10 using ReLU, Swish and ARiA2

| Activation Function | Epoch 25 | Epoch 50 | Epoch 100 |
|---|---|---|---|
| ReLU | 77.96 | 89.37 | 92.55 |
| ARiA2 ($\alpha = 1, \beta = 1$) | 80.94 | 91.06 | 92.02 |
| ARiA2 ($\alpha = 1.25, \beta = 1$) | 82.33 | 92.36 | 93.37 |
| ARiA2 ($\alpha = 1.5, \beta = 1$) | 82.21 | 94.37 | **95.43** |
| ARiA2 ($\alpha = 1.75, \beta = 1$) | 82.87 | 92.92 | 94.01 |
| ARiA2 ($\alpha = 1, \alpha = 2, \beta = 1$) | 80.98 | 91.43 | 92.39 |
| ARiA2 ($\beta = 2$) | 82.73 | 93.97 | 94.99 |
| ARiA2 ($\alpha = 1.25, \beta = 2$) | 83.53 | 91.75 | 92.74 |
| ARiA2 ($\alpha = 1.5, \beta = 2$) | 81.36 | 91.57 | 92.49 |
| ARiA2 ($\alpha = 1.75, \beta = 2$) | 82.82 | 93.95 | 95.00 |
| ARiA2 ($\alpha = 2, \beta = 2$) | 83.63 | 92.49 | 93.41 |

We achieve comparable or better error rate for this model reported in [20], despite training for half the number of epochs by changing the activation function. From Table 4, it can be seen that ARiA2 with $\alpha = 1.5, \beta = 2$ demonstrate highest accuracy.

### 3.1.3 ARiA1

To reduce the computational complexity of optimizing two hyper-parameters in ARiA2, instead, we can employ a simpler version of ARiA with only a single hyper-parameter $\alpha$ (while $\beta$ is retained at 1). As described earlier in section 3, the effect of varying $\alpha$ is more significant than varying $\beta$. This is reflected in our results where, for both models of MNIST, ARiA2 performed better at $\beta = 1$ and $\alpha > 1$. Further, for CIFAR-10 as well, we observe a similar phenomenon where, ARiA2 performs the best for $\alpha = 1.5, \beta = 1$. Only for CIFAR-100, the most superior performance is achieved at $\beta = 2$. However, at $\alpha = 1, \beta = 1$(equivalent to Swish), our model performs better than ReLU. In summary, when higher computational resources are not available, employing ARiA1 is a better choice than employing Swish and/or ReLU and its variants.

Table 4: Testing Accuracy for CIFAR-100 using ReLU, Swish and ARiA2

| Activation Function | Epoch 10 | Epoch 25 | Epoch 50 | Epoch 100 |
|---|---|---|---|---|
| ReLU | 50.92 | 56.59 | 55.67 | 68.06 |
| ARiA2 ($\alpha = 1, \beta = 1$) | 47.18 | 59.48 | 62.51 | 69.02 |
| ARiA2 ($\alpha = 1.5, \beta = 1$) | 48.70 | 58.70 | 62.20 | 68.66 |
| ARiA2 ($\alpha = 1.75, \beta = 1$) | 48.94 | 59.22 | 61.51 | 67.90 |
| ARiA2 ($\alpha = 2, \beta = 1$) | 50.42 | 57.56 | 61.14 | 67.79 |
| ARiA2 ($\alpha = 1, \beta = 2$) | 50.87 | 59.61 | 61.64 | 68.05 |
| ARiA2 ($\alpha = 1.5, \beta = 2$) | 51.96 | 60.19 | 63.55 | **70.17** |
| ARiA2 ($\alpha = 1.75, \beta = 2$) | 51.96 | 59.31 | 61.27 | 67.66 |
| ARiA2 ($\alpha = 2, \beta = 2$) | 52.34 | 57.26 | 60.92 | 67.27 |

## 4  Conclusion

In this work, we introduced and implemented a novel activation function, ARiA, and demonstrated its mathematical relevance, its adaptation to ARiA2 and its comparison to ReLU and Swish activations. The ARiA2 function, that is derived from specialized Richard's Curve provides flexibility to the shape of the function via two hyper-parameters, and thus can be tuned to a particular neural net based task. We illustrated its working on three different datasets, where ARiA2 clearly outperformed ReLU and Swish. Moreover, we also demonstrated that ARiA1 could be a potential option, in cases where optimizing two parameters could be computationally challenging. In summary, the precise control over the non-monotonous convexity is key to ARiA's promising behavior and based on our results it holds the potential to replace ReLU, its variants and Swish in a variety of DNN based prediction tasks.